\documentclass[runningheads]{llncs}
\usepackage[T1]{fontenc}
\usepackage{graphicx}
\usepackage{amsmath,amssymb}
\usepackage{algorithmic}
\usepackage{graphicx}
\usepackage{textcomp}
\usepackage{color}
\usepackage{dutchcal}
\usepackage{multirow}
\usepackage{caption}
\usepackage{hyperref}
\usepackage{url}
\usepackage[most]{tcolorbox}

\usepackage{orcidlink}

\begin{document}

\title{From Images to Insights: Explainable Biodiversity Monitoring with Plain Language Habitat Explanations}

\author{Yutong Zhou\inst{1}\orcidlink{0000-0001-5018-3501} \and
Masahiro Ryo\inst{1,2}\orcidlink{0000-0002-5271-3446}}

\titlerunning{From Images to Insights}

\authorrunning{Y. Zhou and M. Ryo}

\institute{Leibniz Centre for Agricultural Landscape Research (ZALF), Eberswalder Str. 84, 15374, Müncheberg, Germany
\email{yutong.zhou@zalf.de}\\ \and
Brandenburg University of Technology Cottbus–Senftenberg, Platz Der Deutschen Einheit 1, 03046, Cottbus, Germany}

\maketitle              % typeset the header of the contribution

% ----------- Teaser Figure BEFORE Abstract ------------

\begin{center}
\hsize=\textwidth 
\centering
\includegraphics[width=0.66\textwidth,clip]{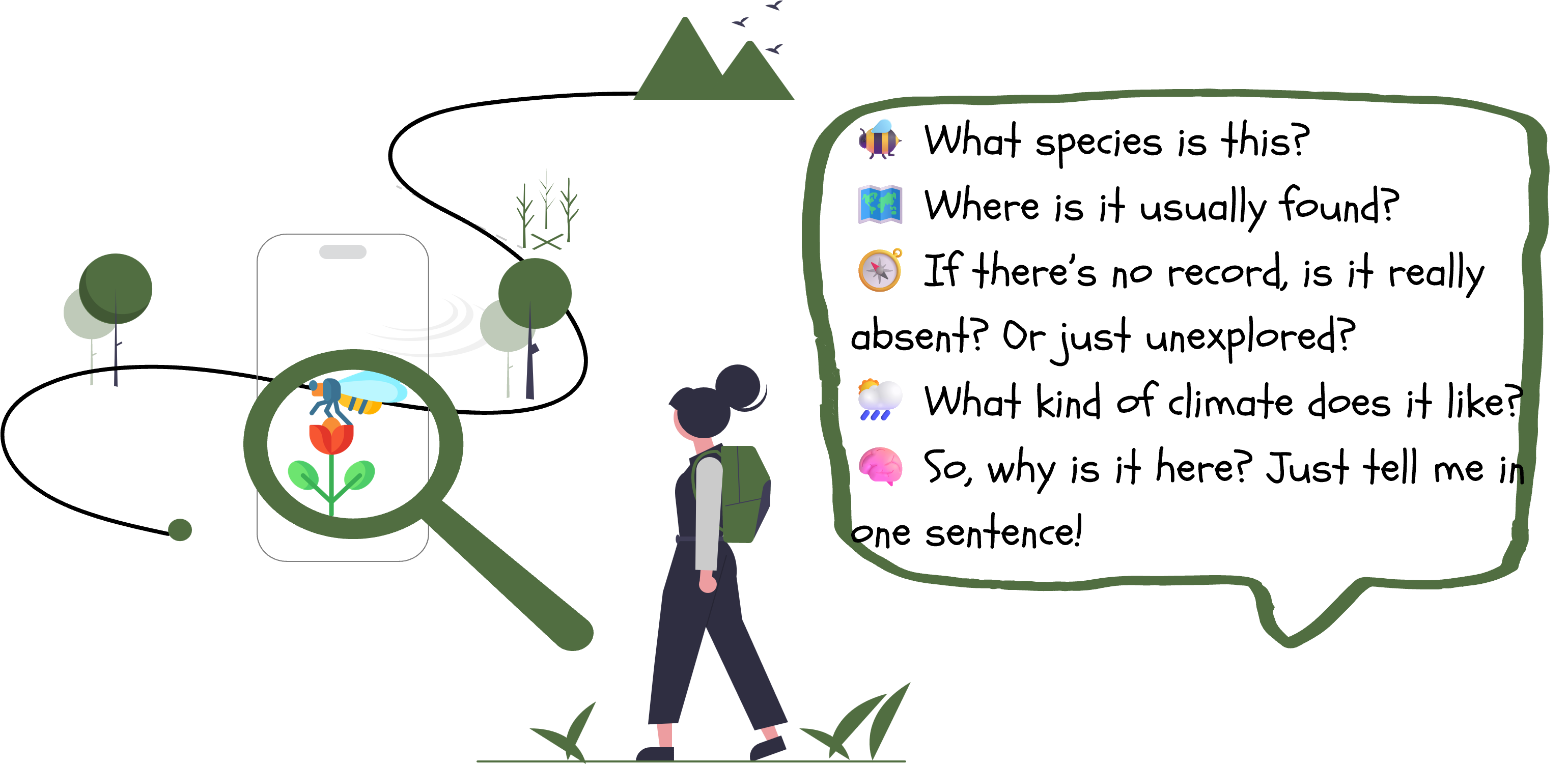}
\captionof{figure}{\textbf{From curiosity to causality.}  
  An explorer encounters bees and flowers in the wild and asks 5 questions. Our system answers them via image-based species identification, species distribution modeling with causal inference, and natural language explanation.}
%Further details setup in sec.\ref{Post-processing}.
\label{Fig1}
\end{center}

\setcounter{footnote}{0}

\begin{abstract}
Explaining why the species lives at a particular location is important for understanding ecological systems and conserving biodiversity. However, existing ecological workflows are fragmented and often inaccessible to non-specialists. We propose an end-to-end visual-to-causal framework that transforms a species image into interpretable causal insights about its habitat preference. The system integrates species recognition, global occurrence retrieval, pseudo-absence sampling, and climate data extraction. We then discover causal structures among environmental features and estimate their influence on species occurrence using modern causal inference methods. Finally, we generate statistically grounded, human-readable causal explanations from structured templates and large language models. We demonstrate the framework on a bee and a flower species and report early results as part of an ongoing project, showing the potential of the multimodal AI assistant backed up by a recommended ecological modeling practice for describing species habitat in human-understandable language. Our code is available at \url{https://github.com/Yutong-Zhou-cv/BioX}.
%Unlike predictive models with limited interpretability, our approach emphasizes explanation and causal reasoning, offering a scalable, transparent tool for ecologists, educators, and policy-makers. 

\end{abstract}
\section{Introduction}
Understanding which environmental factors determine species distributions is a foundational step in ecology, with significant influence for conservation, managing climate risks, and policy. Species distribution modeling (SDM) is a promising tool. However, the workflows require expert knowledge, access to multiple data sources, and fluency with statistical tools, which limits accessibility and only to statistical experts with coding ability. To address these challenges and better answer the questions from Fig. \ref{Fig1}, we propose an end-to-end framework that begins with an input image (\textbf{visual} curiosity) and ends with natural-language ecological insights (\textbf{causal} understanding).

Given a photo, our system automated achieves: (1) identify the species, (2) locate global occurrences via biodiversity APIs, (3) simulate absence background conditions, (4) extract environmental data, (5) learn the causal structure among climatic variables, (6) infer their causal effect on species occurrence, and (7) explain the findings in nature language. Our contribution extends beyond combining existing models, providing a novel multimodal perspective for explainable species distribution modeling and ecological reasoning.

\section{Related Works}

\textbf{Visual Species Recognition.} Existing works in species recognition have achieved high performance. The most notable is BioCLIP\cite{stevens2024bioclip}. Recently, BioCLIP 2\cite{gu2025bioclip} incorporates hierarchical taxonomic embeddings and demonstrates ecological semantic alignment with ecological traits. The iNaturalist Challenge dataset is designed to benchmark the classification and detection of fine-grained species in the wild. Balk \textit{et al.}\cite{balk2024fair} present an automated workflow to extract biological knowledge in the emerging field of imageomics from images.

\noindent\textbf{Species Distribution Modeling (SDM).} Species distribution models (SDMs) are defined as empirical models that aim to estimate the probability of species distributions across spatial and temporal dimensions\cite{wang2025role}. Our framework shifts the focus from correlation-based prediction to causally grounded and interpretable insights. Pseudo-absence generation techniques are proposed to simulate true absences when unavailable. Barbet-Massin \textit{et al.}\cite{barbet2012selecting} explore several strategies for selecting pseudo-absences. %Deep learning-based species distribution models (deep SDMs)\cite{brun2023rank} use geographic predictors and citizen-science data to address sampling gaps.

\noindent\textbf{Causal Graph Learning.} Causal discovery aims to identify directed relationships among variables from data. NOTEARS\cite{zheng2018dags} introduces a continuous optimization framework for learning  Directed Acyclic Graph (DAGs). DAG-GNN\cite{yu2019dag} 
% and GraN-DAG\cite{lachapelle2019gradient} 
extends the framework to nonlinear and variational settings. Santos \textit{et al.}\cite{santos2024causal} conduct a systematic literature review with data-driven causal inference analysis. % Luo \textit{et al.}\cite{luo2024causal} develop a causal discovery model that effectively identifies the causal network within the ecosystem.

\noindent\textbf{Causal Inference.} Causal inference frameworks allow estimation of interventional effects from data. DoWhy\cite{sharma2020dowhy}, a causal inference library for causal graph modeling, quantitatively evaluating causal effects, and validating the causal assumptions. In ecological applications, causal inference can support ecologists to test ecological theory in natural ecosystems from novel data streams \cite{dee2023clarifying}.

\section{Methodology}
We propose an end-to-end interpretable pipeline from species image input to natural-language causal insights about distributional drivers. It is composed of 7 modules as shown in Fig. \ref{Fig2}.

\begin{center}
\hsize=\textwidth 
\centering
\includegraphics[width=\textwidth,clip]{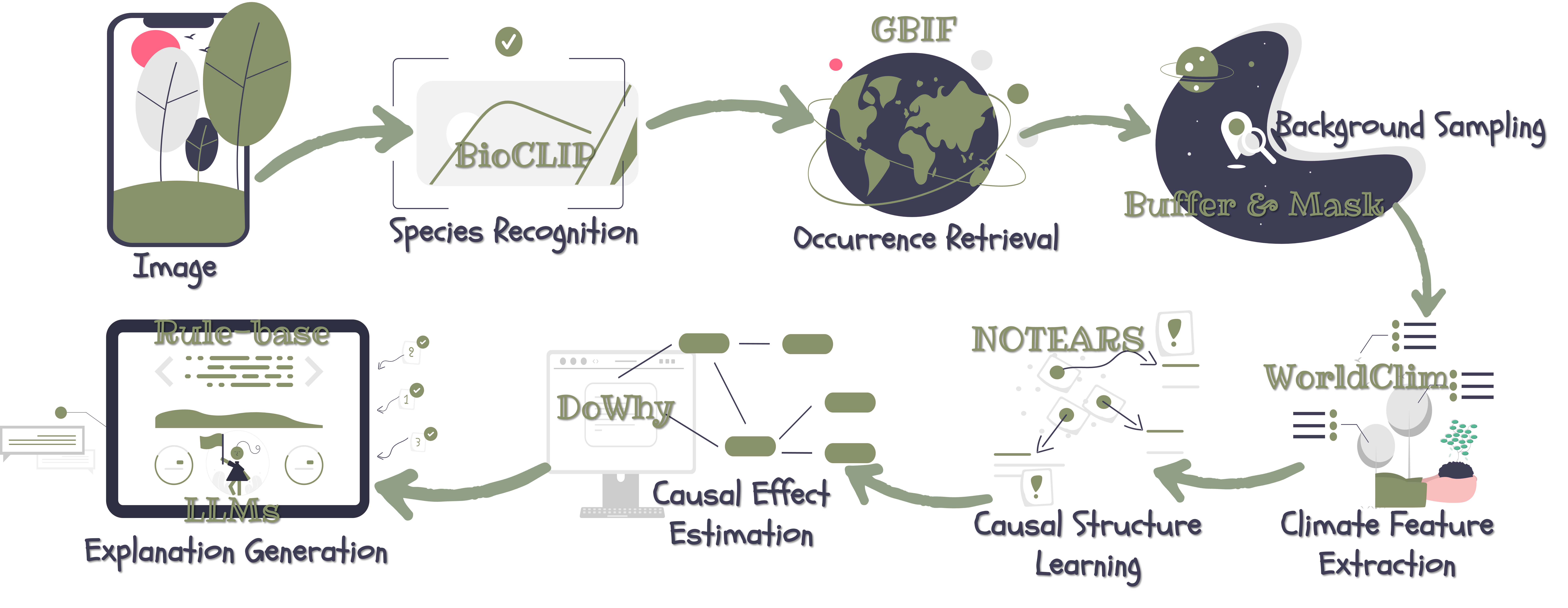}
\captionof{figure}{\textbf{An overview of the visual-to-causal ecological pipeline.}} 
%Given a species image, the system performs species recognition using BioCLIP\cite{stevens2024bioclip}, retrieves occurrence data from GBIF\footnotemark, and generates pseudo-absence points via buffered background sampling\cite{barbet2012selecting}. Climate variables (BIO1–BIO19) are extracted from WorldClim\cite{web:worldclim}. Causal structure among environmental variables is learned using NOTEARS, and DoWhy estimates the causal effects of selected environmental features on species presence. Finally, interpretable explanations are generated using rule-based templates and LLM-based approaches.}
\label{Fig2}
\end{center}

\subsection{Species Recognition}

First, we adopt BioCLIP\cite{stevens2024bioclip}, a vision-language foundation model specifically designed to recognize the species from the input image. BioCLIP is based on the CLIP architecture\cite{radford2021learning} and trained on the TREEOFLIFE-10M dataset, a large-scale structured collection of over 10 million labeled images covering 454,103 unique classes spanning animals, plants, and fungi. Only identifications with a confidence score above 0.80 are retained for the next step. This setting ensures semantic precision and avoids cascading errors in downstream processing.

\subsection{Occurrence Data Retrieval}

Given the predicted species name, we query the Global Biodiversity Information Facility (GBIF)\footnote{\href{https://www.gbif.org/}{GBIF | Global Biodiversity Information Facility: https://www.gbif.org/}}, the largest occurrence database with over 3.1 billion records (accessed June 2025). We retrieve all occurrence records with the following filters:
\begin{itemize}
    \item \textit{basisOfRecord = "HUMAN\_OBSERVATION"}: Only include observations made directly by people, not specimens or sensor data.
    \item \textit{hasCoordinate = TRUE}: Filter for records with valid geographic coordinates (latitude and longitude).
    \item \textit{year $\geq$ 2000}: Limit to recent observations from the year 2000 to present.
    \item \textit{limit = 1000}: Retrieve up to 1000 records per species to balance completeness and efficiency.
\end{itemize}
We extract species names using the GBIF Backbone Taxonomy to account for taxonomic synonymy and ensure consistent querying. The returned data includes species names, latitude, longitude, observation date, and data source. These geo-referenced points are treated as positive samples (Presence = 1).

\subsection{Pseudo-Absence Sampling}
True absence data is rarely available in ecological datasets. To construct a binary classification framework (Presence $\in$ {0,1}), we improve the pseudo-absence sampling strategy inspired by \cite{barbet2012selecting}: (1) Expand the bounding box by ±1° latitude and longitude ($\sim$111 km): Create a spatial buffered region around occurrences to define the sampling area. (2) Equally sample 2$\times$ points within the buffered region: Randomly generate background points without spatial bias. (3) Filter out points within 5 km of a presence record: To avoid mislabeling the likely presence of species as absence. (4) Mask out marine areas using Natural Earth land polygons\footnote{\href{https://www.naturalearthdata.com/downloads/10m-physical-vectors/10m-land/}{Natural Earth | Land 1:10m: https://www.naturalearthdata.com/downloads/10m-physical-vectors/10m-land/}}: Ensure that background points fall only on terrestrial land. These pseudo-absence points are treated as negative samples (Presence = 0).
%This approach aims to reduce the inclusion of false absences while covering plausible climatic space.

\subsection{Environmental Feature Extraction}
For each point (presence and pseudo-absence), we extract 19 bioclimatic variables (BIO1–BIO19) from WorldClim v2.1\cite{fick2017worldclim} at a spatial resolution of 2.5 arc-minutes ($\sim$5 km). These variables capture ecologically meaningful climate dimensions for species distribution modeling techniques, including: Temperature metrics, such as BIO1 (Annual mean temperature), BIO5 (Max temperature of warmest month), and Precipitation metrics, such as BIO12 (Annual precipitation), BIO15 (Precipitation seasonality).
%We spatially index the coordinates and extract the raster values into a tabular form.

\subsection{Environmental Causal Graph Learning}
Climate variables are not independent. Therefore, understanding the underlying causal structure is crucial before estimating the effects on species' occurrence. We explore NOTEARS\cite{zheng2018dags} to learn the causal structure among the 19 climate variables. Specifically, we apply both the linear NOTEARS and nonlinear variant (NOTEARS MLP) to capture different types of dependencies, including linear trends and more complex nonlinear interactions. NOTEARS formulates structure learning as a continuous optimization problem that solves for a weighted adjacency matrix \textit{W} with a differentiable acyclicity constraint \(\min _W \frac{1}{2 n}\|X-X W\|_F^2+\lambda\|W\|_1, \quad \text {subject to } h(W)=\operatorname{tr}\left(e^{W \circ W}\right)-d=0\), to efficiently get directed acyclic graphs (DAGs) from observational data. \textit{X} is the matrix of environmental variables, $\lambda$ is the hyperparameter that controls the strength of L1 regularization, \textit{h(W)} enforces the DAG constraint, $\circ$ is element-wise product, \textit{d} is the number of nodes.
%Compared to traditional combinatorial approaches, NOTEARS offers scalability and robustness, making it well-suited for modeling complex dependencies in environmental systems where variables (e.g., temperature and precipitation metrics) are often correlated yet causally distinct.

\subsection{Causal Inference on Species Occurrence}
% While traditional SDMs can discover correlations between environmental variables and species presence, they cannot answer complex policy-relevant questions.
%such as: What would happen to species X if temperature increases by 2°C? To address this, 
We adopt a causal inference approach using DoWhy\cite{sharma2020dowhy}, which allows us to estimate the causal effect of climate features on species occurrence. Our method approximates real-world interventions using structural causal models (SCMs) grounded in the previously learned DAG from NOTEARS.

Specifically, for each of the top-5 candidate variables (treatments) identified from the causal graph, we define (1) Treatment: a bioclimatic variable (\textit{e.g.}, BIO4: temperature seasonality); (2) Outcome: species presence (binary label); (3) Confounder set: automatically selected via the backdoor criterion using the NOTEARS DAG.
We use DoWhy to define the causal graph and verify identifiability. Then, estimate the Average Treatment Effect (ATE) using stratified propensity score adjustment. 
%Retain only statistically significant estimates ($p \textless 0.05$) for downstream explanation. 
This integration of graph-driven backdoor adjustment with modern causal estimators represents a key novelty of our framework. 
% It allows us to generate insights that are not only statistically sound but also biologically interpretable and intervention-relevant—supporting use cases in conservation planning, ecological forecasting, and climate resilience assessment.
%The framework is also adaptable to evolving environmental baselines: as climate change shifts global conditions, both the causal graph and intervention estimates can be re-evaluated on updated climate scenarios, enabling long-term ecological monitoring.

\subsection{Explanation Generation}
Finally, we generate causal explanations using two complementary approaches: (1) \textbf{Rule-based generation}, following the template ( ``BIO'' is bioclimatic variable name, ``SP.'' is species name), such as, If ATE $\geq$ 0.1, generate ``\textit{High {BIO} strongly promotes {SP.} presence. This likely reflects a core habitat requirement.}''; If -0.05 $\textless$ ATE $\textless$ 0, generate ``\textit{{BIO} has a weak negative effect.}''. (2) \textbf{LLM-enhanced generation}, using LLAMA3.3-70B\cite{meta_llama3.3} with constrained prompting. To ensure scientific rigor, prompts are designed to ensure factual grounding (such as \textit{“Ensure the explanation is realistic, grounded in ecological reasoning, and free from vague generalizations”}, \textit{“Write 1 sentence explaining the most likely ecological mechanism behind the causal influence”}). The output is designed for environmental scientists, conservation policy experts, or education platforms requiring accessible scientific communication.

\section{Experimental Results}

\subsection{Pilot Study Setup}

To assess the effectiveness and interpretability of our framework, we conduct experiments on two representative species from different ecological niches as pilot studies: \textit{Osmia parietina (Western mason bee)}, a wild European bee that nests in sunny mountain habitats, and \textit{Ajuga reptans (bugle)}, a ground-covering blue spring flower that thrives in moist conditions.
Data retrieved via GBIF for each species\cite{bee,flower} is $\sim$1,000 presence points, with 2,000 background points. The final data structure is: \textit{[Latitude, Longitude, BIO1, ..., BIO19, Presence (0/1)]}.

% \subsection{Causal Analysis}
% We learn environmental DAGs using NOTEARS. For Osmia parietina, the learned graph revealed strong dependencies that align with known ecological theory: 

% “BIO10 (Mean Temperature of Warmest Quarter) → BIO1 (Annual Mean Temp) 

% BIO15 (Precipitation Seasonality) → BIO14 (Precipitation of Driest Month)

% BIO2 (Mean Diurnal Range) → BIO7(Temperature Annual Range)”

\subsection{Causal Analysis and Example Explanations}
An example of Ajuga reptans, the learned environmental DAGs revealed strong dependencies, such as BIO 11 (Mean Temperature of Coldest Quarter) → BIO6 (Min Temperature of Coldest Month), BIO 10 (Mean Temperature of Warmest Quarter) → BIO7 (Temperature Annual Range). Then DoWhy estimated statistically ATEs = 0.13 for BIO11 and ATEs = -0.03 for BIO10. This suggests that its presence is positively influenced by milder winters. However, extreme temperature fluctuations limit its distribution and abundance.
%suggesting that the species is sensitive to heat stress, drought, and temperature variability. 

Table \ref{tab:explanation_comparison} compares highlight insights of rule-based explanations (from feature importance scores) with LLM-generated ecological interpretations for each species. The LLM outputs provide more nuanced, biologically grounded reasoning, connecting variable effects to thermoregulation, drought sensitivity, and habitat preferences. The detailed results are omitted due to the page limitation. 

\vspace{-8mm}  
\begin{table}
    \centering
    \footnotesize
    \caption{Comparison between rule-based and LLM-based ecological explanations.}
    \renewcommand{\arraystretch}{1.3}
    \begin{tabular}{|p{2cm}|p{5cm}|p{5cm}|}
        \hline
        \textbf{Species} & \textbf{Rule-based Explanation} & \textbf{LLM-based Explanation} \\
        \hline
        \textbf{Osmia parietina} & 
        Higher \textbf{BIO2} (Mean Diurnal Range) moderately suppresses presence, suggesting sensitivity to daily temperature fluctuation. \textbf{BIO10} (Mean Temp of Warmest Quarter) has a weak positive effect. & 
        The species prefers \textbf{warm, moist, and thermally stable} environments. It is negatively affected by \textbf{high temperature variability}, especially \textbf{large diurnal ranges} and \textbf{seasonal extremes}, reflecting sensitivity to thermal stress and limits in thermoregulation. \\
        \hline
        \textbf{Ajuga reptans} & 
        \textbf{BIO11} (Mean Temp of Coldest Quarter) strongly promotes presence. \textbf{BIO2} moderately suppresses, and \textbf{BIO5} (Max Temp of Warmest Month) imposes a strong negative constraint. & 
        A temperate species favoring \textbf{mild winters} and \textbf{stable precipitation}. \textbf{Heat stress, drought}, and \textbf{temperature variability} can reduce presence and abundance, indicating vulnerability to climate extremes despite general adaptability to moist environments. \\
        \hline
    \end{tabular}
    \label{tab:explanation_comparison}
\end{table}
\vspace{-8mm}

\section{Conclusion}
% We introduced a novel visual-to-causal, end-to-end framework that transforms a species image into interpretable causal insights about its habitat in plain language. Our approach integrates species recognition, geospatial data retrieval, climate variable extraction, causal graph discovery, and causal effect estimation into a single automated pipeline. This provides a novel perspective to answer why a species is found where it is, grounding predictions in interpretable ecological mechanisms rather than black-box patterns. 

% This work is still at a preliminary stage. Only two case studies were included, so broader generalization remains to be validated. The ecological validity of generated explanations has not yet been systematically evaluated through user or expert studies. These limitations highlight directions for future research.

% Although the model currently captures only a limited subset of relevant variables, 
% %By combining computer vision, geospatial ecology, causal inference, and language generation, our framework provides a scalable and interpretable prototype for biodiversity monitoring. 
% it shows significant potential in conservation planning, climate risk assessment, and ecological education, and provides a foundation for future AI systems that reason about the natural world in transparent and human-aligned ways. 

We introduced a novel visual-to-causal end-to-end framework that transforms species images into causal ecological explanations. By integrating species recognition, geospatial retrieval, climate variable extraction, causal graph learning, and inference, the pipeline provides an interpretable alternative to correlation-based distribution modeling and answers why species occur where they do. 

The study remains preliminary, as only two case studies were presented, and the ecological validity of the explanations has not yet been systematically assessed. Despite these limitations, the approach shows strong potential for conservation planning, climate risk assessment, and ecological education, and provides a foundation for future AI systems that reason about the natural world in transparent and human-aligned ways. 

\begin{credits}
\subsubsection{\ackname} The project on which this report is based was funded by the German Federal Ministry of Education and Research within the Research Initiative for the Conservation of Biodiversity (FEdA) under the funding code 16LW0682K. The responsibility for the content of this publication lies with the author. We thank the data publishers who made their datasets available through GBIF.org and whose contributions are listed on the respective download DOI pages.
\end{credits}

\bibliographystyle{splncs04}
\bibliography{ref}

\end{document}